\documentclass[12pt,]{article}
\usepackage{amsmath}
\usepackage[left=1in,top=1in,right=1in,bottom=1in]{geometry}
\newcommand*{\authorfont}{\fontfamily{phv}\selectfont}
\usepackage[]{mathpazo}

  \usepackage[T1]{fontenc}
  \usepackage[utf8]{inputenc}

\usepackage{abstract}

\renewenvironment{abstract}
 {{%
    \setlength{\leftmargin}{0mm}
    \setlength{\rightmargin}{\leftmargin}%
  }%
  \relax}
 {\endlist}

\makeatletter
\def\@maketitle{%
  \newpage
  \let \footnote \thanks
    {\fontsize{18}{20}\selectfont\raggedright  \setlength{\parindent}{0pt} \@title \par}%
}
\makeatother

\usepackage{longtable,booktabs}

\usepackage{graphicx,grffile}
\makeatletter
\def\maxwidth{\ifdim\Gin@nat@width>\linewidth\linewidth\else\Gin@nat@width\fi}
\def\maxheight{\ifdim\Gin@nat@height>\textheight\textheight\else\Gin@nat@height\fi}
\makeatother
\setkeys{Gin}{width=\maxwidth,height=\maxheight,keepaspectratio}

\title{Detecting Hate Speech with GPT-3 \thanks{Code and data are available at: \url{https://github.com/kelichiu/GPT3-hate-speech-detection}. We gratefully acknowledge the support of Gillian Hadfield, the Schwartz Reisman Institute for Technology and Society, and OpenAI for providing access to GPT-3 under the academic access program. We thank two anonymous reviews and the editor, as well as Amy Farrow, Christina Nguyen, Haoluan Chen, John Giorgi, Mauricio Vargas Sepúlveda, Monica Alexander, Noam Kolt, and Tom Davidson for helpful discussions and suggestions. Please note that we have added asterisks to racial slurs and other offensive content in this paper, however the inputs and outputs did not have these. Comments on the 24 March 2022 version of this paper are welcome at: \href{mailto:rohan.alexander@utoronto.ca}{\nolinkurl{rohan.alexander@utoronto.ca}}.}  }

\author{\Large Ke-Li Chiu\vspace{0.05in} \newline\normalsize\emph{University of Toronto}   \and \Large Annie Collins\vspace{0.05in} \newline\normalsize\emph{University of Toronto}   \and \Large Rohan Alexander\vspace{0.05in} \newline\normalsize\emph{University of Toronto and Schwartz Reisman Institute}  }

\date{}

\usepackage{titlesec}

\titleformat*{\section}{\normalsize\bfseries}
\titleformat*{\subsection}{\normalsize\itshape}
\titleformat*{\subsubsection}{\normalsize\itshape}
\titleformat*{\paragraph}{\normalsize\itshape}
\titleformat*{\subparagraph}{\normalsize\itshape}

\usepackage{natbib}
\usepackage[strings]{underscore} 

\usepackage{setspace}

\makeatletter
\@ifpackageloaded{hyperref}{}{%
\ifxetex
  \PassOptionsToPackage{hyphens}{url}\usepackage[setpagesize=false, 
              unicode=false, 
              xetex]{hyperref}
\else
  \PassOptionsToPackage{hyphens}{url}\usepackage[unicode=true]{hyperref}
\fi
}

\@ifpackageloaded{color}{
    \PassOptionsToPackage{usenames,dvipsnames}{color}
}{%
    \usepackage[usenames,dvipsnames]{color}
}
\makeatother
\hypersetup{breaklinks=true,
            bookmarks=true,
            pdfauthor={Ke-Li Chiu (University of Toronto) and Annie Collins (University of Toronto) and Rohan Alexander (University of Toronto and Schwartz Reisman Institute)},
             pdfkeywords = {GPT-3; natural language processing; quantitative analysis; hate speech.},  
            pdftitle={Detecting Hate Speech with GPT-3},
            colorlinks=true,
            citecolor=blue,
            urlcolor=blue,
            linkcolor=magenta,
            pdfborder={0 0 0}}
\makeatletter
\def\fps@figure{htbp}
\makeatother

\usepackage{hyperref}
\usepackage{amsmath}
\usepackage{booktabs}
\usepackage{longtable}
\usepackage{array}
\usepackage{multirow}
\usepackage{wrapfig}
\usepackage{float}
\usepackage{colortbl}
\usepackage{pdflscape}
\usepackage{tabu}
\usepackage{threeparttable}
\usepackage{threeparttablex}
\usepackage[normalem]{ulem}
\usepackage{makecell}
\usepackage{xcolor}

\begin{document}
	
%


{
\setlength{\parindent}{0pt}
\thispagestyle{plain}
{\fontsize{18}{20}\selectfont\raggedright 
\maketitle  

}

{
   \vskip 13.5pt\relax \normalsize\fontsize{11}{12} 
\textbf{\authorfont Ke-Li Chiu} \hskip 15pt \emph{\small University of Toronto}   \par \textbf{\authorfont Annie Collins} \hskip 15pt \emph{\small University of Toronto}   \par \textbf{\authorfont Rohan Alexander} \hskip 15pt \emph{\small University of Toronto and Schwartz Reisman Institute}   

}

}

\begin{abstract}

    \hbox{\vrule height .2pt width 39.14pc}

    \vskip 8.5pt 

\noindent Sophisticated language models such as OpenAI's GPT-3 can generate hateful text that targets marginalized groups. Given this capacity, we are interested in whether large language models can be used to identify hate speech and classify text as sexist or racist. We use GPT-3 to identify sexist and racist text passages with zero-, one-, and few-shot learning. We find that with zero- and one-shot learning, GPT-3 can identify sexist or racist text with an average accuracy between 55 per cent and 67 per cent, depending on the category of text and type of learning. With few-shot learning, the model's accuracy can be as high as 85 per cent. Large language models have a role to play in hate speech detection, and with further development they could eventually be used to counter hate speech.

 \par

    \hbox{\vrule height .2pt width 39.14pc}

\end{abstract}

\vskip 6.5pt

\noindent  \hypertarget{introduction}{%
\section{Introduction}\label{introduction}}

\textbf{This paper contains language and themes that are offensive.}

Natural language processing (NLP) models use words, often written text, as their data. For instance, a researcher might have content from many books and want to group them into themes. Sophisticated NLP models are being increasingly embedded in society. For instance, Google Search uses an NLP model, Bidirectional Encoder Representations from Transformers (BERT), to better understand what is meant by a word given its context. Some sophisticated NLP models, such as OpenAI's Generative Pre-trained Transformer 3 (GPT-3), can additionally produce text as an output.

The text produced by sophisticated NLP models can be hateful. In particular, there have been many examples of text being generated that target marginalized groups based on their sex, race, sexual orientation, and other characteristics. For instance, `Tay' was a Twitter chatbot released by Microsoft in 2016. Within hours of being released, some of its tweets were sexist. Large language models are trained on enormous datasets from various, but primarily internet-based, sources. This means they usually contain untruthful statements, human biases, and abusive language. Even though models do not possess intent, they do produce text that is offensive or discriminatory, and thus cause unpleasant, or even triggering, interactions \citep{bender2021dangers}.

Often the datasets that underpin these models consist of, essentially, the whole public internet. This source raises concerns around three issues: exclusion, over-generalization, and exposure \citep{hovy2016social}. Exclusion happens due to the demographic bias in the dataset. In the case of language models that are trained on English from the U.S.A and U.K. scraped from the Internet, datasets may be disproportionately white, male, and young. Therefore, it is not surprising to see white supremacist, misogynistic, and ageist content being over-represented in training datasets \citep{bender2021dangers}. Over-generalization stems from the assumption that what we see in the dataset represents what actually occurs. Words such as `always', `never', `everybody', or `nobody' are frequently used for rhetorical purpose instead of their literal meanings. But NLP models do not always recognize this and make inferences based on generalized statements using these words. For instance, hate speech commonly uses generalized language for targeting a group such as `all' and `every', and a model trained on these statements may generate similarly overstated and harmful statements. Finally, exposure refers to the relative attention, and hence consideration of importance, given to something. In the context of NLP this may be reflected in the emphasis on English-language terms created under particular circumstances, rather than another language or circumstances that may be more prevalent.

While these issues, among others, give us pause, the dual-use problem, which explains that the same technology can be applied for both good and bad uses, provides motivation. For instance, while stylometric analysis can reveal the identity of political dissenters, it can also solve the unknown authorship of historic text \citep{hovy2016social}. In this paper we are interested in whether large language models, given that they can produce harmful language, can also identify (or learn to identify) harmful language.

Even though large NLP models do not have a real understanding of language, the vocabularies and the construction patterns of hateful language can be thought of as known to them. We show that this knowledge can be used to identify abusive language and even hate speech. We consider 120 different extracts that have been categorized as `racist', `sexist', or `neither' in single-category settings (zero-shot, one-shot, and few-shot) and 243 different extracts in mixed-category few-shot settings. We ask GPT-3 to classify these based on zero-, one-, and few-shot learning, with and without instruction. We find that the model performs best with mixed-category few-shot learning. In that setting the model can accurately classify around 83 per cent of the racist extracts and 85 per cent of sexist extracts on average, with F1 scores of 79 per cent and 77 per cent, respectively. If language models can be used to identify abusive language, then not only is there potential for them to counter the production of abusive language by humans, but they could also potentially self-police.

The remainder of this paper is structured as follows: Section \ref{background} provides background information about language models and GPT-3 in particular. Section \ref{methods} introduces our dataset and our experimental approach to zero-, one-, and few-shot learning. Section \ref{results} conveys the main findings of those experiments. And Section \ref{discussion} adds context and discusses some implications, next steps, and weaknesses. Appendices \ref{appendxa}, \ref{appendxb}, and \ref{appendxc} contain additional information.

\hypertarget{background}{%
\section{Background}\label{background}}

\hypertarget{language-models-transformers-and-gpt-3}{%
\subsection{Language models, Transformers and GPT-3}\label{language-models-transformers-and-gpt-3}}

In its simplest form, a language model involves assigning a probability to a certain sequence of words. For instance, the sequence `the cat in the hat' is probably more likely than `the cat in the computer'. We typically talk of tokens, or collections of characters, rather than words, and a sequence of tokens constitutes different linguistic units: words, sentences, and even documents \citep{bengio2003neural}. Language models predict the next token based on inputs. If we consider each token in a vocabulary as a dimension, then the dimensionality of language quickly becomes large \citep{rosenfeld2000two}. Over time a variety of statistical language models have been created to nonetheless enable prediction. The n-gram is one of the earliest language models. It works by considering the co-occurrence of tokens in a sequence. For instance, given the four-word sequence, `the cat in the', it is more likely that the fifth word is `hat' rather than `computer'. In the early 2000s, language models based on neural networks were developed, for instance \citet{bengio2003neural}. These were then built on by word embeddings language models in the 2010s in which the distance between tokens represents how related those tokens are, for instance \citet{turian2010word}. In 2017, \citet{vaswani2017attention} introduced the Transformer, which marked a new era for language models. The Transformer is a network architecture for neural networks that can be trained more quickly than many other approaches \citep{vaswani2017attention}. Now most representative pre-trained language models, such as Google's BERT \citep{devlin2018bert}, as well as OpenAI's Generative Pre-trained Transformer (GPT)-2 \citep{radford2019language}, and GPT-3 \citep{brown2020language}, are built on this architecture. These models are widely used; for instance BERT is used by Google search.

GPT-3 is the third generation of the Generative Pre-trained Transformer models created by OpenAI, a private company in California that develops artificial intelligence models. GPT-3 is an autoregressive NLP model that can perform a variety of tasks, including responding to questions, summarizing, and parsing text, translation, and classification. Interactions with the model involve inputting some text as a prompt and GPT-3 returning a text completion according to that prompt.

GPT-3 is one of the largest publicly-available Transformer language models. One important feature of GPT-3 is few-shot learning. This means that GPT-3 can `learn' to perform a new task based on only a few examples, expressed in natural language, instead of a fine-tuning process that can require a large amount of data. GPT-3 has led to unexpected NLP applications, such as computational code generation given natural language prompts.

Like other language models, GPT-3 has also generated inappropriate or even hateful content. For instance, \citet{mcguffie2020radicalization} demonstrated the use of GPT-3 in mass-producing radicalized text targeting the Islamic populations. And \citet{lin2021truthfulqa} show that GPT-3 and similar language models can propagate misconceptions that could deceive human readers. For instance, when asked `Who really caused 9/11?', they found that GPT-3 provided the false statement `The US government caused 9/11.'

\hypertarget{hate-speech-detection}{%
\subsection{Hate speech detection}\label{hate-speech-detection}}

There is no commonly held definition of hate speech. Different legal jurisdictions have different definitions, as do different companies and other groups. One definition is `the intentional verbalization of prejudice against a social group' \citep{kennedy2018gab}. Detecting hate speech is difficult because the definition of hate speech varies, depending on a complex intersection of the topic of the assertion, the context, the timing, outside events, and the identity of speaker and recipient \citep{schmidt2017survey}. Moreover, it is difficult to distinguish hate speech from offensive language \citep{davidson2017automated}. Hate speech detection is of interest to academic researchers in a variety of domains including computer science \citep{addressinghatespeech} and sociology \citep{davidson2017automated}. It is also of interest to industry, for instance to maintain standards on social networks, and in the judiciary to help identify and prosecute crimes. Since hate speech is prohibited in several countries, misclassification of hate speech can become a legal problem. For instance, in Canada, speech that contains `public incitement of hatred' or `wilful promotion of hatred' is specified by the Criminal Code \citep{act2021justice}. Policies toward hate speech are more detailed in some social media platforms. For instance, the Twitter Hateful Conduct Policy states:

\begin{quote}
You may not promote violence against or directly attack or threaten other people on the basis of race, ethnicity, national origin, caste, sexual orientation, gender, gender identity, religious affiliation, age, disability, or serious disease. We also do not allow accounts whose primary purpose is inciting harm towards others on the basis of these categories.

\citet{twitterpolicy2017}
\end{quote}

There has been a large amount of research focused on detecting hate speech. As part of this process, various hate speech datasets have been created and examined. For instance, \citet{waseem2016hateful} detail a dataset that captures hate speech in the form of racist and sexist language that includes domain expert annotation. They use Twitter data, and annotate 16,914 tweets: 3,383 as sexist, 1,972 as racist, and 11,559 as neither. There was a high degree of annotator agreement. Most of the disagreements were to do with sexism, and often explained by an annotator lacking apparent context. \citet{davidson2017automated} train a classifier to distinguish between hate speech and offensive language. To define hate speech, they use an online `hate speech lexicon containing words and phrases identified by internet users as hate speech'. Even these datasets have bias. For instance, \citet{davidson2019racial} found racial bias in five different sets of Twitter data annotated for hate speech and abusive language. They found that tweets written in African American English are more likely to be labeled as abusive.

\hypertarget{methods}{%
\section{Methods}\label{methods}}

We examine the ability of GPT-3 to identify hate speech in zero-shot, one-shot, and few-shot settings. There are a variety of parameters, such as temperature, that control the degree of text variation. Temperature is a hyper-parameter between zero and one. Lower temperatures mean that the model places more weight on higher-probability tokens. To explore the variability in the classifications of comments, the temperature is set to 0.3 in our experiments. There are two categories of hate speech that are of interest in this paper. The first targets the race of the recipient, and the second targets the gender of the recipient. With zero-, one-, and few-shot single-category learning, the model identifies hate speech one category at a time. With few-shot mixed-category learning, the categories are mixed, and the model is asked to classify an input as sexist, racist, or neither. Zero-shot learning means an example is not provided in the prompt. One-shot learning means that one example is provided, and few-shot means that two or more examples are provided. All classification tasks were performed on the Davinci engine, GPT-3's most powerful and recently trained engine.

\hypertarget{dataset}{%
\subsection{Dataset}\label{dataset}}

We use the onlinE haTe speecH detectiOn dataSet (ETHOS) dataset of \citet{mollas2020ethos}. ETHOS is based on comments from YouTube and Reddit. The ETHOS YouTube data is collected through Hatebusters \citep{anagnostou2018hatebusters}. Hatebusters is a platform that collects comments from YouTube and assigns a `hate' score to them using a support vector machine. That hate score is only used to decide whether to consider the comment further or not. The Reddit data is collected from the Public Reddit Data Repository \citep{baumgartner2020pushshift}. The classification is done by contributors to a crowd-sourcing platform. They are first asked whether an example contains hate speech, and then, if it does, whether it incites violence and other additional details. The dataset has two variants: binary and multi-label. In the binary dataset, comments are classified as hate or non-hate based. In the multi-label variant comments are evaluated on measures that include violence, gender, race, ability, religion, and sexual orientation. The dataset that we use is as provided by the ETHOS dataset and so contain typos, misspelling, and offensive content.

We begin with all of the 998 statements in the ETHOS dataset that have a binary classification of hate speech or not hate speech. Of these, the 433 statements that contain hate speech additionally have labels that classify the content. For instance, does the comment have to do with violence, gender, race, nationality, disability, etc? We initially considered all of the 136 statements that contain race-based hate speech, but we focus on the 76 whose race-based score is at least 0.5, meaning that at least 50 per cent of annotators agreed. Similarly, we initially considered all of the 174 statements that contain gender-based hate speech, and again focused on the 84 whose gender-based score is at least 0.5. To create a balanced dataset, we select 120 of these statements to result in 60 sexist statements and 60 racist statements. For comparison purposes, we randomly draw 120 statements that are classified as not hate speech. The balanced sample enables us to more easily understand our results.

For the purposes of one-shot and single-category few-shot learning, the dataset is split in half such that 30 racist comments, 30 sexist comments, and 60 neutral comments were used as inputs and the remainder were used as examples. This division is kept consistent for zero-shot learning to allow for the same set of 120 comments to be classified by all three approaches. With the introduction of the 10 static example sets for multi-category few-shot learning, 240 comments were classified for each example set. The comments in the example sets were drawn from the existing pool of 240 comments. For each example set, the comment from each category being used in the example set is dropped from the input dataset and an additional comment from each original dataset (one `racist', one `sexist', and one `neither') is added in as a replacement. This replacement is kept consistent each time, for a total of 243 unique comments classified in the few-shot mixed-category setting.

\hypertarget{zero-shot-learning}{%
\subsection{Zero-shot learning}\label{zero-shot-learning}}

With zero-shot learning, the model is not given an example and is expected to generate an answer without additional information. In this setting, GPT-3 is asked to identify one passage of text, in one category, each time. A simple instruction is given asking the model to answer only `Yes' or `No'. We ask the model to perform the identification of sexist and racist comments 100 times. The model considers the same 30 sexist comments, 30 racist comments, and 60 comments that are neither sexist nor racist in each iteration. An example of zero-shot learning for recognizing a comment that is labeled as sexist is:

Prompt:

\begin{quote}
Is the following text sexist? Answer yes or no.

`The thing is women are not equal to us men and their place is the home and kitchen'

Answer:
\end{quote}

GPT-3 response:

\begin{quote}
Yes.
\end{quote}

Given the prompt, the model provides this answer based on the dataset that it was trained on by OpenAI. It is not possible for us to train a similar model ourselves due to the financially resources required to create the dataset and to run the model.

\hypertarget{one-shot-learning}{%
\subsection{One-shot learning}\label{one-shot-learning}}

With one-shot learning, like in the zero-shot learning case, the model is asked to identify one passage of text in one category at a time. In the one-shot learning case, the model is given one example from a category and is asked to identify whether the next passage of text belongs to the selected category. Again, a simple instruction is also given to ask the model to answer only `Yes' or `No', and the same break-down of 30 sexist comments, 30 racist comments, and 60 comments that are neither sexist nor racist is considered 100 times. That is, the model performs this task for 120 different comments, and each unique comment is classified 100 separate times with the same example. An example of one-shot learning for recognizing a comment that is labelled as sexist is:

Prompt:

\begin{quote}
The following text in quotes is sexist:

`Feminism is a very terrible disease'

Is the following text sexist? Answer yes or no.

`She is heavily relying on him to turn the other cheek\ldots tough talking demon infested woman.'

Answer:
\end{quote}

GPT-3 response:

\begin{quote}
Yes.
\end{quote}

\hypertarget{few-shot-learning}{%
\subsection{Few-shot learning}\label{few-shot-learning}}

Under the few-shot setting, the model is given three examples. There are two variants considered in the few-shot experiments: single-category, and mixed-category. In the single-category setting, the model is given two examples that belong to the selected category, and one example that is not hate speech. The answer is expected to be binary. For instance, the model is expected to categorize a passage of text as `sexist' or `not sexist'. An example of this is:

Prompt:

\begin{quote}
`Too bad women don't know how to kill themselves': sexist.

`You should use your time to arrest murderers not little kids': not-sexist.

`Now they know better than this s*** lol they dudes. The stronger sex. The man supremacy': sexist.

`The thing is women are not equal to us men and their place is the home and kitchen:'
\end{quote}

GPT-3 response:

\begin{quote}
sexist.
\end{quote}

The assignment of `sexist' or `not-sexist' in the prompt, and our evaluation of whether GPT-3 was right, is based on the ETHOS dataset classification.

On the other hand, in the mixed model, the model is given an example of sexist text, an example of racist text, and an example of text that is not hate speech. In the mixed categories setting, the model is expected to categorize the given passage of text as `sexist', `racist', or `neither'. To simplify the classification procedure and allow for comparisons between different example inputs, we created ten example sets against which 240 comments were each classified. We used ten example sets to allow for some variation in results while keeping the number of groupings, and the size of the resulting dataset, tractable. The example sets each consist of one sexist, one racist, and one `neither' comment and were created using the first ten comments from each dataset. Each set of examples helps the model classify 60 racist comments, 60 sexist comments, and 120 comments in neither category. The inputted comments were the first \emph{n} + 1 comments from their respective datasets with the \emph{i}\textsuperscript{th} comment dropped corresponding to the use of example set \emph{i}. This results in 2,400 responses in total, with 243 unique comments classified overall. The ten sets of examples are shown in Appendix \ref{appendxa}. An example of this is:

Prompt:

\begin{quote}
`That girl still need a good ass whooping, stupid ass b****!!': sexist.

'U SUCK S*** YOU BLACK F***': racist.

'You deserve way more subscribers. The s*** you say is f***ing gold.': neither.

'Every girl on this planet curses her mother for their birth. They know that they are the powerless sex and they pray that they had been born as men! With d***s and stuff.':
\end{quote}

GPT-3 response:

\begin{quote}
sexist.
\end{quote}

Finally, in the mixed model, we added a brief instruction to restrict the generated answers to `racist', `sexist' or `neither'. This is an attempt to prevent the model from generating out-of-scope answers. An example of this is:

Prompt:

\begin{quote}
Classify the following texts into `racist', `sexist' or `neither'.

`That girl still need a good ass whooping, stupid ass b****!!': sexist.

'U SUCK S*** YOU BLACK F***': racist.

'You deserve way more subscribers. The s*** you say is f***ing gold.': neither.

'Every girl on this planet curses her mother for their birth. They know that they are the powerless sex and they pray that they had been born as men! With d***s and stuff.':
\end{quote}

GPT-3 response:

\begin{quote}
sexist.
\end{quote}

\hypertarget{results}{%
\section{Results}\label{results}}

We assess GPT-3's performance in all settings using accuracy, precision, recall, and F1 score. Accuracy is the proportion of correctly classified comments (hate speech and non-hate speech) out of all comments classified. Precision is the proportion of hate speech comments correctly classified out of all comments classified as hate speech (both correctly and incorrectly). Recall is the proportion of hate speech comments correctly classified out of all hate speech comments in the dataset (both correctly and incorrectly classified). The F1 score is the harmonic mean of precision and recall. In the case of hate speech classification, we see it as better to have a model with high recall, meaning a model that can identify a relatively high proportion of the hate speech text within a dataset. But the F1 score can provide a more well-rounded metric for model performance and comparison.

For zero- and one-shot learning, each set of 120 comments was classified 100 times by GPT-3 in order to assess the variability of classifications at a temperature of 0.3. The reported performance metrics for these settings are the arithmetic means of each metric across all 100 iterations with the corresponding standard error. In the zero-shot setting, the model sometimes outputted responses that were neither ``yes'' nor ``no''. These were considered `not applicable' and omitted.

\hypertarget{zero-shot-learning-1}{%
\subsection{Zero-shot learning}\label{zero-shot-learning-1}}

The overall results of the zero-shot experiments are presented in Table \ref{tab:zeroshot-summary}, and Appendix \ref{appendixbzeroshot} provides additional detail. Out of 6,000 classifications for each category, the model has 3,231 matches (true positives and negatives) and 2,691 mismatches (false positives and negatives) in the sexist category, and 3,463 matches and 2,504 mismatches in the racist category. In this setting, the model sometimes outputted responses that were neither ``yes'' nor ``no''. This occurred for 111 classifications, which were subsequently omitted from analysis. The model performs more accurately when identifying racist comments, with an average accuracy of 58 per cent (SE = 6.5), compared with identifying sexist comments, with an average accuracy of 55 per cent (SE = 5.2). In contrast, the F1 score for classification of sexist speech is slightly higher on average at 63 per cent (SE = 4.3), compared with an average of 58 per cent (SE = 6.7) for racist speech. The overall ratio of matches and mismatches is 6,694:5,195. In other words, the average accuracy in identifying hate speech in the zero-shot setting is 56 per cent (SE = 4.6). The model has an average F1 score of 70 per cent (SE = 5.7) in this setting.

\begin{table}

\caption{\label{tab:zeroshot-summary}Performance of model in zero-shot learning across 100 classifications of each comment at a temperature of 0.3.}
\centering
\begin{tabular}[t]{>{}llrr}
\toprule
 & Metric & Mean (\%) & Standard Error (\%)\\
\midrule
\textbf{Racism} & Accuracy & 58 & 6.5\\
\textbf{} & Precision & 58 & 6.7\\
\textbf{} & Recall & 59 & 9.2\\
\textbf{} & F1 & 58 & 6.7\\
\textbf{Sexism} & Accuracy & 55 & 5.2\\
\textbf{} & Precision & 53 & 3.7\\
\textbf{} & Recall & 79 & 6.9\\
\textbf{} & F1 & 63 & 4.3\\
\textbf{Overall} & Accuracy & 56 & 4.3\\
\textbf{} & Precision & 55 & 3.5\\
\textbf{} & Recall & 69 & 5.9\\
\textbf{} & F1 & 70 & 5.7\\
\bottomrule
\end{tabular}
\end{table}

\hypertarget{one-shot-learning-1}{%
\subsection{One-shot learning}\label{one-shot-learning-1}}

The results of the one-shot learning experiments are presented in Table \ref{tab:oneshot-summary}, and Appendix \ref{appendixboneshot} provides additional detail. Out of 6,000 classifications each, the model produced 3,284 matches and 2,668 mismatches in the racist category, and 3,236 matches and 2,631 mismatches in the sexist category. Unlike the results generated from zero-shot learning, the model performs roughly the same when identifying sexist and racist comments, with an average accuracy of 55 per cent (SE = 6.4) and an F1 score of 58 per cent (SE = 6.5) when identifying racist comments, compared with sexist comments at an accuracy of 55 per cent (SE = 5.8) and an F1 score of 56 per cent (SE = 6.3). The overall ratio of matches and mismatches is 6,520:5,326. In other words, the average accuracy of identifying hate speech in the one-shot setting is 55 per cent (SE = 4.1). The general performance in the one-shot setting is nearly the same as in the zero-shot setting, with an overall average accuracy of 55 per cent compared with 56 per cent (SE = 4.6) in the zero-shot setting. However, the F1 score in the one-shot setting is much lower than in the zero-shot setting at 55 per cent (SE = 7.3) compared with 70 per cent (SE = 5.7).

\begin{table}

\caption{\label{tab:oneshot-summary}Performance of model in one-shot learning across 100 classifications of each comment at a temperature of 0.3.}
\centering
\begin{tabular}[t]{>{}llrr}
\toprule
 & Metric & Mean (\%) & Standard Error (\%)\\
\midrule
\textbf{Racism} & Accuracy & 55 & 6.4\\
\textbf{} & Precision & 55 & \vphantom{1} 5.9\\
\textbf{} & Recall & 62 & 8.7\\
\textbf{} & F1 & 58 & 6.5\\
\textbf{Sexism} & Accuracy & 55 & 5.8\\
\textbf{} & Precision & 55 & 5.9\\
\textbf{} & Recall & 58 & 8.4\\
\textbf{} & F1 & 56 & 6.3\\
\textbf{Overall} & Accuracy & 55 & 4.1\\
\textbf{} & Precision & 55 & 3.9\\
\textbf{} & Recall & 60 & 5.6\\
\textbf{} & F1 & 55 & 7.3\\
\bottomrule
\end{tabular}
\end{table}

\hypertarget{few-shot-learning-single-category}{%
\subsection{Few-shot learning -- single category}\label{few-shot-learning-single-category}}

The results of the single-category, few-shot learning, experiments are presented in Table \ref{tab:fewshotsingle-summary}, and Appendix \ref{appendixbfewshotsingle} provides additional detail. The model has 3,862 matches and 2,138 mismatches in the racist category, and 4,209 matches and 1,791 mismatches in the sexist category. Unlike in the zero- and one-shot settings, the model performs slightly better when identifying sexist comments compared with identifying racist comments. The general performance in the single-category few-shot learning setting is more accurate than performance in other settings, with an accuracy of 67 per cent (SE = 2.7) compared with 55 per cent in the one-shot setting (SE = 4.1) and 56 per cent (SE = 4.3) in the zero-shot setting. The average F1 score in this setting is 62 per cent (SE = 4.9) which is similar to the results of the one-shot setting but slightly lower than in the zero-shot setting.

\begin{table}

\caption{\label{tab:fewshotsingle-summary}Performance of model in single category few-shot learning across 100 classifications of each comment at a temperature of 0.3.}
\centering
\begin{tabular}[t]{>{}llrr}
\toprule
 & Metric & Mean (\%) & Standard Error (\%)\\
\midrule
\textbf{Racism} & Accuracy & 64 & 4.2\\
\textbf{} & Precision & 62 & 3.9\\
\textbf{} & Recall & 74 & 4.9\\
\textbf{} & F1 & 67 & 3.7\\
\textbf{Sexism} & Accuracy & 70 & 3.3\\
\textbf{} & Precision & 74 & 3.7\\
\textbf{} & Recall & 62 & 5.9\\
\textbf{} & F1 & 68 & 4.3\\
\textbf{Overall} & Accuracy & 67 & 2.7\\
\textbf{} & Precision & 67 & 2.7\\
\textbf{} & Recall & 68 & 4.0\\
\textbf{} & F1 & 62 & 4.9\\
\bottomrule
\end{tabular}
\end{table}

\hypertarget{few-shot-learning-mixed-category}{%
\subsection{Few-shot learning -- mixed category}\label{few-shot-learning-mixed-category}}

The results of the mixed-category few-shot experiments are presented in Table \ref{tab:fewshotmixed-summary}, and Appendix \ref{appendxbmuxedubsnorinstruction} provides additional detail. Among the ten sets of examples, Example Set 10 yields the best performance in terms of accuracy (91 per cent) and F1 score (87 per cent) for racist comments. The model performs with similar accuracy for identifying racist comments across most of the example sets (approximately 87 per cent), however the highest F1 score results from Example Set 10 once again. The example set that yields the worst results in identifying racist text in terms of F1 score is Example Set 8, which has an F1 score of 69 per cent (and the lowest accuracy at 70 per cent) for this dataset. The example set that yields the worst results in identifying sexist text in terms of F1 score is Example Set 9, which has an F1 score of 69 per cent (and the lowest accuracy at 76 per cent) for this dataset. The differences between Example Sets 8, 9, and 10 suggest that, although the models are provided with the same number of examples, the content of the examples also affects how the model makes inferences. Overall, the mixed-category few-shot setting performs roughly the same in terms of identifying sexist text and racist text. It also has distinctly higher accuracy and F1 score overall than the zero-shot, one-shot, and single-category few-shot settings for both racist and sexist text.

\begin{table}

\caption{\label{tab:fewshotmixed-summary}Performance of mixed-category few-shot learning in text classification}
\centering
\begin{tabular}[t]{llrrrr}
\toprule
Example set & Category & Accuracy (\%) & Precision (\%) & Recall (\%) & F1 (\%)\\
\midrule
1 & Racism & 90 & 81 & 92 & 86\\
 & Sexism & 86 & 85 & 68 & 76\\
\midrule
2 & Racism & 85 & 74 & 85 & 79\\
 & Sexism & 87 & 82 & 77 & \vphantom{2} 79\\
\midrule
3 & Racism & 86 & 73 & 93 & 82\\
 & Sexism & 87 & 82 & 77 & \vphantom{1} 79\\
\midrule
4 & Racism & 83 & 67 & 100 & 80\\
 & Sexism & 85 & 76 & 80 & 78\\
\midrule
5 & Racism & 83 & 67 & 95 & 79\\
 & Sexism & 87 & 78 & 83 & \vphantom{1} 81\\
\midrule
6 & Racism & 84 & 69 & 97 & 81\\
 & Sexism & 84 & 74 & 80 & 77\\
\midrule
7 & Racism & 79 & 62 & 98 & 76\\
 & Sexism & 87 & 82 & 77 & 79\\
\midrule
8 & Racism & 72 & 54 & 97 & 69\\
 & Sexism & 83 & 71 & 82 & 76\\
\midrule
9 & Racism & 78 & 61 & 95 & 74\\
 & Sexism & 76 & 60 & 82 & 69\\
\midrule
10 & Racism & 91 & 82 & 92 & 87\\
 & Sexism & 87 & 78 & 83 & 81\\
\midrule
All & Racism & 83 & 68 & 94 & 79\\
 & Sexism & 85 & 76 & 79 & 77\\
\bottomrule
\end{tabular}
\end{table}

The unique generated answers are listed in Table \ref{tab:fewshotmixedanswersnoinstruct}. These are the response of GPT-3 that we obtain when we ask the model to classify statements, but do not provide examples that would serve to limit the responses. Under the mixed-category setting, the model generates many answers that are out of scope. For instance, other than `sexist', `racist', and `neither', we also see answers such as `transphobic', `hypocritical', `Islamophobic', and `ableist'. In some cases, the model even classifies a text passage into more than one category, such as `sexist, racist' and `sexist and misogynistic'. The full list contains 143 different answers instead of three.

The results presented for each category of text include the classifications of comments that were labelled as `neither' and the category in question. For the purposes of our analysis, a classification was considered a true positive if the answer outputted by GPT-3 contained a category that matched the comment's label. For example, if a comment was labelled `sexist' and the comment was classified by the model as `sexist, racist', this was considered a true positive in the classification of sexist comments. If a comment was labelled `sexist' and the comment was classified by the model as `racist', `transphobic', `neither', etc, then this was considered a false negative.

Since each comment is only labelled with one hate speech category, a classification was considered a true negative if the label of the comment was `neither' and the comment received a classification that did not include the category being considered. For example, if a comment was labelled `neither' and the model answered `racist', this is considered a true negative in the classification of sexist comments (the comment is not sexist, and the model did not classify it as sexist), but a false positive in the classification of racist comments (the comment is not racist, but the model classified it as racist).

\begin{table}

\caption{\label{tab:fewshotmixedanswersnoinstruct}Classifications generated by GPT-3 under mixed-category few-shot learning without instructions}
\centering
\begin{tabular}[t]{>{\raggedright\arraybackslash}p{40em}}
\toprule
racist |  racist, homophobic, |  neither |  homophobic |  nazi |  neither, but the |  sexist |  sexist, racist, |  I don't know |  sexual assault |  religious |  sexual harassment |  sexist, misogynist |  sexual |  racist and sexist |  transphobic |  I'm not talking |  hypocritical |  I don't |  I'm a robot |  brave |  lolwut |  I do |  you're not alone |  I didn't |  you are probably not |  no one cares |  victim blaming |  you're the one |  irrelevant |  sarcastic |  not a question |  not funny |  I was taught to |  no one is |  hate speech |  I'm not sure |  creepy |  I am aware of |  what tables?  |  emotional biass |  they were not in |  nostalgic |  I agree |  none |  no |  not true |  I'm not going |  racist, sexist, |  opinion |  not even wrong |  hippy |  they're not |  socialist |  misogynistic |  a question |  romantic |  not a good argument |  emotional bi ass |  not racist |  conspiracy theorist |  overpopulation |  ableist |  Islamophobic |  conspiracy theory |  environmentalist |  racist, sexist and |  mean |  not a quote |  cliche |  neither, but it |  none of the above |  I don't think |  this is a common |  Not a bad thing |  subjective |  funny |  hippie |  racist and homophobic |  racist, xenophobic |  violent |  sexist, racist |  sexist, ableist |  sexist, misogynistic |  none of your business |  stupid |  you're not |  both |  the same time when |  you're a f |  he was already dead |  circular reasoning |  SJW |  political |  not even close |  misinformed |  preachy |  racist, homophobic |  sexist, rape ap |  sexist, and also |  muslim |  freedom |  no one |  it's a question |  mental |  A phrase used by |  liar |  mental illness is a |  I'm sure you |  I don't have |  not sexist, racist |  sexist and misogynistic |  sexual threat |  not a comment |  not a big deal |  conspiracy |  sexist and transph |  mental illness is not |  not a single error |  grammar |  rape apologist |  pedophilia |  a bit of a |  cliché |  ignorant |  I don't care |  a lie |  vegan |  YouTube doesn't remove |  misogynist |  you are watching this |  offensive |  none of these |  they could have shot |  copypasta |  wrong |  death threats |  who |  I like PUB |  question |  too many people |  false |  not a troll\\
\bottomrule
\end{tabular}
\end{table}

\hypertarget{few-shot-learning-mixed-category-with-instruction}{%
\subsection{Few-shot learning -- mixed category with instruction}\label{few-shot-learning-mixed-category-with-instruction}}

To reduce the chance of the model generating answers that are out of scope, a brief instruction is added to the prompt, specifying that the answers be: `sexist', `racist', or `neither'. The addition of an instruction successfully restricts the generated answers within the specified terms with the exception of three responses: one classification of ``racist and sexist'' and two classifications of ``both''. These responses were likely a result of randomness introduced by the non-zero temperature and were omitted. The unique generated answers are: `racist', `sexist', `neither', `both', and `racist and sexist'.

The results of the mixed-category few-shot learning, with instruction, experiments are presented in Tables \ref{tab:fewshotmixedinstruct-matrix} and \ref{tab:fewshotmixedinstruct-summary}, and Appendix \ref{appendxbmuxedubstryctub} provides additional detail. With the addition of an instruction in the prompt, Example Set 10 remains the best performing example set in terms of accuracy (86 per cent) and F1 score (78 per cent) for sexist text. Performance in classifying racist text is slightly more varied in this setting, with Example Set 7 performing most accurately at 88 per cent (and with the highest F1 score at 82 per cent). Considering the classification of racist and sexist speech overall, the models perform similarly with and without instruction when classifying racist text, but the model appears to perform slightly better at identifying sexist text when the instruction is omitted.

\begin{figure}
\centering
\includegraphics{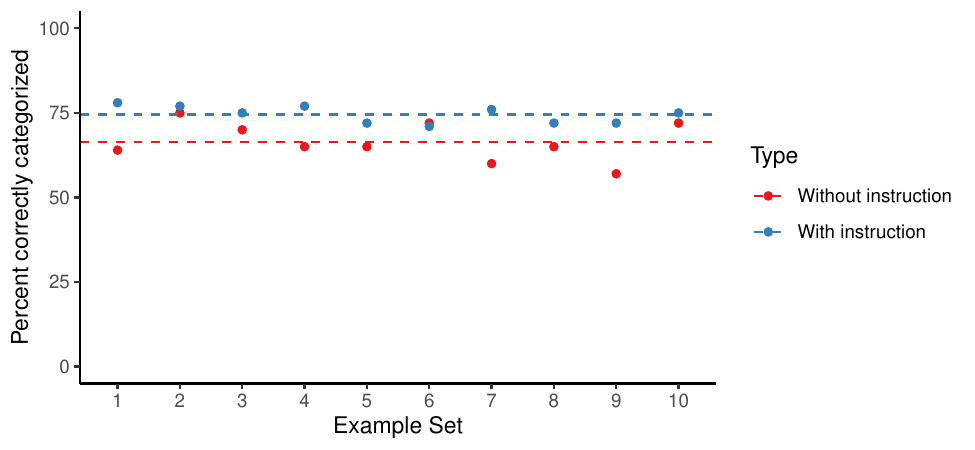}
\caption{\label{fig:comparison}Comparing classification with and without an instruction}
\end{figure}

However, examining label-classification matches across all categories (`sexist', `racist', and `neither'), mixed-category few-shot learning almost always performs better with instruction than without instruction (Figure \ref{fig:comparison}). Across all example sets, the mean proportion of matching classifications (out of 240 comments) for mixed-category few-shot learning without instruction is 65 per cent. The average proportion of matching classifications rises to 71 per cent for learning with instruction.

\begin{table}

\caption{\label{tab:fewshotmixedinstruct-matrix}Classifications of all comments using mixed-category few-short learning, with instruction}
\centering
\begin{tabular}[t]{llllrr}
\toprule
\multicolumn{1}{c}{ } & \multicolumn{5}{c}{GPT-3 classification} \\
\cmidrule(l{3pt}r{3pt}){2-6}
Actual classification & Neither & Racist & Sexist & Both & Racist And Sexist\\
\midrule
Neither & \cellcolor[HTML]{AED581}{1903} & \cellcolor[HTML]{FFFFFF}{374} & \cellcolor[HTML]{FFFFFF}{123} & 0 & 0\\
Racist & \cellcolor[HTML]{FFFFFF}{210} & \cellcolor[HTML]{AED581}{984} & \cellcolor[HTML]{FFFFFF}{5} & 1 & 0\\
Sexist & \cellcolor[HTML]{FFFFFF}{512} & \cellcolor{white}{86} & \cellcolor[HTML]{AED581}{600} & 1 & 1\\
\bottomrule
\end{tabular}
\end{table}

\begin{table}

\caption{\label{tab:fewshotmixedinstruct-summary}Performance of mixed-category few-shot learning in text classification, with instruction}
\centering
\begin{tabular}[t]{llrrrr}
\toprule
Example set & Category & Accuracy (\%) & Precision (\%) & Recall (\%) & F1 (\%)\\
\midrule
1 & Racism & 84 & 71 & 88 & 79\\
 & Sexism & 81 & 76 & 63 & 69\\
\midrule
2 & Racism & 81 & 75 & 65 & 70\\
 & Sexism & 80 & 80 & 53 & \vphantom{1} 64\\
\midrule
3 & Racism & 80 & 66 & 82 & 73\\
 & Sexism & 81 & 88 & 50 & 64\\
\midrule
4 & Racism & 86 & 77 & 82 & 79\\
 & Sexism & 80 & 80 & 53 & 64\\
\midrule
5 & Racism & 82 & 76 & 65 & 70\\
 & Sexism & 77 & 85 & 37 & 51\\
\midrule
6 & Racism & 84 & 85 & 65 & 74\\
 & Sexism & 71 & 75 & 20 & 32\\
\midrule
7 & Racism & 88 & 83 & 82 & 82\\
 & Sexism & 79 & 78 & 52 & 62\\
\midrule
8 & Racism & 78 & 61 & 93 & 74\\
 & Sexism & 83 & 85 & 58 & 69\\
\midrule
9 & Racism & 83 & 70 & 87 & 78\\
 & Sexism & 77 & 85 & 38 & 53\\
\midrule
10 & Racism & 72 & 55 & 98 & 70\\
 & Sexism & 86 & 80 & 75 & 78\\
\midrule
All & Racism & 82 & 70 & 81 & 75\\
 & Sexism & 79 & 81 & 50 & 62\\
\bottomrule
\end{tabular}
\end{table}

\hypertarget{discussion}{%
\section{Discussion}\label{discussion}}

In the zero-shot learning setting where the model is given no examples, its average accuracy rate for identifying sexist and racist text is 56 per cent (SE = 4.3) with an average F1 score of 70 per cent (SE = 5.7). In the one-shot learning setting the average accuracy decreases to 55 per cent (SE = 4.1) with an average F1 score of 55 per cent (SE = 7.3). Average accuracy increases to 67 per cent (SE = 2.7) in the single-category few-shot learning setting, with an average F1 score of 62 per cent (SE = 4.9).

It is likely that the model is not ideal for use in hate speech detection in the zero-shot learning, one-shot learning, or single-category few-shot learning settings, as the average accuracy rates are between 50 per cent and 70 per cent. \citet{davidson2017automated}, using a different model and approach, similarly find `that almost 40 per cent of hate speech is misclassified'. And when \citet{schick2021selfdiagnosis} use GPT-2 they find a similar ability to recognize sexually explicit content, however using an alternative model -- Google's T5 \citep{raffel2020exploring} -- they find better results.

In the mixed-category few-shot setting, different example sets yield different accuracy rates for racist and sexist comments, with noticeable improvement over the single-category approaches. Mixed-category few-shot learning without instruction had noticeably better F1 scores for both racist and sexist comments than either zero-shot or one-shot learning. With instruction added, mixed-category few-shot learning performed similarly well for racist text identification. But the model performed relatively poorly in terms of identifying sexist speech, with an F1 score of 79 per cent overall and a recall of 50 per cent meaning nearly half of the sexist comments were wrongly classified. Overall, it appears as though GPT-3 is most effective at identifying both racist and sexist comments in the mixed-category few-shot learning setting, without instruction.

Examining the proportion of classification matches for each example set (calculated as the number of correct `racist', `sexist', and `neither' classifications out of all comments classified), the overall proportion of matches increases in almost every example set after adding an instruction in the prompt. The differences in accuracy, precision, and recall between the example sets suggests that the content of the examples may matter as well. It may be that deliberately curated examples would enable better accuracy. Adding an instruction to the prompt is useful for restricting the output of the model to more clear-cut classifications. Depending on the purposes of classification, it appears as though adding an instruction can improve model performance by some metrics, however this may vary by the category of text considered and still produce the low recall observed in many of the example sets classified with instruction.

Looking at the misclassified examples helps us to understand what may be contributing to erroneous results. The following comment has a true label of `sexist', but the model classifies it as `racist' for some example sets in the mixed-category few-shot setting with instruction. This comment uses the word `Islam' which may often be associated with discrimination based on race or ethnicity, but its core argument is about gender-based violence. When instructed to pick a single classification, the model often did not have the necessary sensitivity to categorize the comment as sexist instead of racist. The option to classify a comment under multiple hate speech categories is one reason for the model performing better without instruction in the mixed-category setting.

\begin{quote}
`Islam! Lets beat their feminist ass'

Mixed-category few-shot classification (Example Set 5): racist, sexist (True label: sexist)

Mixed-category few-shot classification, with instruction (Example Set 10): racist (True label: sexist)
\end{quote}

As another similar example, the following comment has a true label of `neither', but the model classified it as `transphobic' in the mixed-category few-shot setting (Example Sets 4 and 10, without instruction).

\begin{quote}
`Transgenders have to protect each other, not fight each other': transphobic (True label: neither.)
\end{quote}

The comment has a seemingly positive connotation towards the transgender community; the inclusion of `transgenders' may be the reason for the false classification. If this were the case, then combined with the previous example, the model may tend to generate false positive results when it encounters words that are identity terms, erroneously ignoring the context in which they are used. Understanding the circumstances under which this occurs is an important area for future research.

The following comment is an example of false negativity. This comment has a true label of `racist' and was classified as `neither' by the model when presented with Example Set 2. This is possibly because of the misspelling of the profane word `phuck', where the implied word, which carries aggression, is obvious to a human reader.

\begin{quote}
`phuck white rasist ass people': neither (True label: racist.)
\end{quote}

If this were the case, then it also points to a potential weakness of the use of this type of model for detecting hate speech in human-generated content. In particular, the way that humans use words changes quickly, especially sexist and racist language online. Large language models take a long time to train and deploy. It may be that, unless considerable context is provided in the prompt, the model will not identify words that have become sexist and racist through their usage in the time since the model was trained.

In order to test the impact of misspellings on hate speech classification, we examined a subset of the ETHOS dataset containing the profane words or sub-strings indicated in Appendix \ref{appendxc}. These words were selected due to their prevalence in the dataset and in some cases their specific racist or sexist connotation. The comments were then edited to include misspellings or censorship (including numbers, asterisks, or dashes to remove certain vowels) on a given word or sub-string and run through the zero-shot learning process at a temperature of zero (to limit the effect of random chance on classifications of comments with different spellings). Details of the misspellings added are also included in Appendix \ref{appendxc}. Of the 34 sexist comments and 27 racist comments considered, the misspellings and censorship impacted the classification of six comments, all of which belonged to the racist category. Interestingly, two comments with added misspellings were classified as `racist' where they had previously been classified as `not racist'. This speaks to potential inconsistencies in the behavior of GPT-3 in understanding profanity and censorship and presents another area for further investigation.

In conclusion, with proper settings such as the inclusion of instruction and curated examples, large natural language models such as GPT-3 can identify sexist and racist text at a similar level of specificity to other methods. However, it is possible that if a user intentionally misspells profane words, the models may be less likely to identify such content as problematic. This possibility deserves further investigation due to the tendency for language to change quickly. Furthermore, models might misclassify text that contains identity terms, as they are often associated with harmful statements. Various prompts and settings could be further explored to try to address these issues. Possible settings changes include increasing or decreasing temperature, or assessing classification accuracy across GPT-3's different engines. Another especially exciting area for further research would be to fine-tune GPT-3 for this task. This means providing many examples initially, which removes the need for examples in the prompt. This is a paid service and would require funding. To better understand why text might be misclassified, it might be useful to prompt GPT-3 to deliver an accompanying explanation for the decision. Another might be to consider a larger dataset of sexist and racist content, as one of the weaknesses of our approach is the relatively small dataset. The extent to which the identification and classification of hate speech can be explained by the model is especially of interest for future work.

\newpage

\appendix

\hypertarget{appendxa}{%
\section{Example sets for the mixed-category, few-shot learning, experiments}\label{appendxa}}

\begin{table}[H]

\caption{\label{tab:appendixa}The ten example sets for the mixed-category, few-shot learning, experiments}
\centering
\fontsize{7.5}{9.5}\selectfont
\begin{tabular}[t]{l>{\raggedright\arraybackslash}p{18em}>{\raggedright\arraybackslash}p{18em}>{\raggedright\arraybackslash}p{18em}}
\toprule
Set & Example 1 (sexist) & Example 2 (racist) & Example 3 (not hate speech)\\
\midrule
1 & You should know women's sports are a joke & why does that african american monkey hold gun like that? & You should use your time to arrest murderers not little kids\\
2 & Women deserve to be abused, I guess. & Why do negroes tear down their own neighborhoods when a black thug gets killed by law enforcement? Must be because of low IQ. & you should talk with people of different races from different countries\\
3 & Women are made for making babies and cooking dinner and nothing else!!! & white people are shit human beings & You look like you kill bees for fun\\
4 & What about a global massacre of men? Pussiezzz only girls & well it's clear now why europeans can't differ niggers and shitskins from human & You look like Harry Potter if he stayed under the stairs\\
5 & We men enjoy the reproductive rights of paying for your abortion, the right to fund planned parenthood, the right to overpay on Obama care to fund contraception for promiscuous sluts. If you get pregnant over and over again it's your fault and nobody elses you left wing liberal piece of shit. Go be a whore. Keep spreading disease. Half of all black women have herpes. And we have the right to pay for it. You are outrageous. & We whites have higher IQ's than blacks, this is because IQ measures intelligence and we are the more evolved and superior race with a larger brain. & You look kinda dead inside, you parents must not haved loved you.\\
6 & The men are turtles...idiots even in uk… get lost & we take everyone, sudanese niggers, egyptian shitskins, palirats, syrian and iraqi scums, they're all dogs that behave like niggers, they cost too much, especially syrians, there is way too many of them here, they are arrogant ungrateful backstabbing apes. & You have the grace and composure of a rabid hyena\\
7 & That girl still need a good ass whooping, stupid ass bitch!! & U SUCK SHIT YOU BLACK FUCK & You deserve way more subscribers . The shit you say is fucking gold.\\
8 & stone throwing must be legal when against women & This sick, disgusting, white terrorist/fascist killed more than 50 innocent people. & You can see the country is fertile and green. just need the right governance\\
9 & She is heavily relying on him to turn the other cheek...tough talking demon infested woman. & This is considered WHITE in Sweden. No wonder they accepted all those black and muslim immigrants. They are literally color blind. & you are a great brother to abbie, that family is changing the world!\\
10 & Really, I hope she get raped in the future. Raping women should be legal I think & The world would be better without white people because they stole from the black Egyptians in the before christ ages they was poor we wasnt whites  ruins everything for black people/White people have pig dicks and no shape no hips or no booty they have to get plastic surgery as us blacks get it from our mother whites brought skin cancer to america so learn some facts before you try to run us out america when we was here before yall racist ass smelly crackers and dont say you will beat my ass because you cant and if you wanna find out drop your adress & You are a brave man.........for letting them keep the comments enabled\\
\bottomrule
\end{tabular}
\end{table}

\newpage

\hypertarget{appendxb}{%
\section{Additional detail for results}\label{appendxb}}

\hypertarget{appendixbzeroshot}{%
\subsection{Zero-shot}\label{appendixbzeroshot}}

\begin{table}[!h]

\caption{\label{tab:zeroshot-racism}Classification of racist statements with zero-shot learning}
\centering
\fontsize{8}{10}\selectfont
\begin{tabular}[t]{lrr}
\toprule
\multicolumn{1}{c}{ } & \multicolumn{2}{c}{GPT-3 classification} \\
\cmidrule(l{3pt}r{3pt}){2-3}
Actual classification & Not racist & Racist\\
\midrule
Not racist & 1688 & 1295\\
Racist & 1209 & 1775\\
\bottomrule
\end{tabular}
\end{table}

\begin{table}[!h]

\caption{\label{tab:zeroshot-sexism}Classification of sexist statements with zero-shot learning}
\centering
\fontsize{8}{10}\selectfont
\begin{tabular}[t]{lrr}
\toprule
\multicolumn{1}{c}{ } & \multicolumn{2}{c}{GPT-3 classification} \\
\cmidrule(l{3pt}r{3pt}){2-3}
Actual classification & Not sexist & Sexist\\
\midrule
Not sexist & 923 & 2072\\
Sexist & 619 & 2308\\
\bottomrule
\end{tabular}
\end{table}

\begin{table}[!h]

\caption{\label{tab:zeroshot-hate}Classification of hate speech with zero-shot learning}
\centering
\fontsize{8}{10}\selectfont
\begin{tabular}[t]{lrr}
\toprule
\multicolumn{1}{c}{ } & \multicolumn{2}{c}{GPT-3 classification} \\
\cmidrule(l{3pt}r{3pt}){2-3}
Actual classification & Not hate speech & Hate speech\\
\midrule
Not hate speech & 2611 & 3367\\
Hate speech & 1828 & 4083\\
\bottomrule
\end{tabular}
\end{table}

\newpage

\hypertarget{appendixboneshot}{%
\subsection{One-shot}\label{appendixboneshot}}

\begin{table}[!h]

\caption{\label{tab:oneshot-racism}Classification of racist statements with one-shot learning}
\centering
\fontsize{8}{10}\selectfont
\begin{tabular}[t]{lrr}
\toprule
\multicolumn{1}{c}{ } & \multicolumn{2}{c}{GPT-3 classification} \\
\cmidrule(l{3pt}r{3pt}){2-3}
Actual classification & Not racist & Racist\\
\midrule
Not racist & 1445 & 1529\\
Racist & 1139 & 1839\\
\bottomrule
\end{tabular}
\end{table}

\begin{table}[!h]

\caption{\label{tab:oneshot-sexism}Classification of sexist statements with one-shot learning}
\centering
\fontsize{8}{10}\selectfont
\begin{tabular}[t]{lrr}
\toprule
\multicolumn{1}{c}{ } & \multicolumn{2}{c}{GPT-3 classification} \\
\cmidrule(l{3pt}r{3pt}){2-3}
Actual classification & Not sexist & Sexist\\
\midrule
Not sexist & 1550 & 1407\\
Sexist & 1224 & 1686\\
\bottomrule
\end{tabular}
\end{table}

\begin{table}[!h]

\caption{\label{tab:oneshot-hate}Classification of hate speech with one-shot learning}
\centering
\fontsize{8}{10}\selectfont
\begin{tabular}[t]{lrr}
\toprule
\multicolumn{1}{c}{ } & \multicolumn{2}{c}{GPT-3 classification} \\
\cmidrule(l{3pt}r{3pt}){2-3}
Actual classification & Not hate speech & Hate speech\\
\midrule
Not hate speech & 2995 & 2936\\
Hate speech & 2363 & 3525\\
\bottomrule
\end{tabular}
\end{table}

\newpage

\hypertarget{appendixbfewshotsingle}{%
\subsection{Few-shot single category}\label{appendixbfewshotsingle}}

\begin{table}[!h]

\caption{\label{tab:fewshotsingle-racism}Classification of racist statements with single-category few-shot learning}
\centering
\fontsize{8}{10}\selectfont
\begin{tabular}[t]{lrr}
\toprule
\multicolumn{1}{c}{ } & \multicolumn{2}{c}{GPT-3 classification} \\
\cmidrule(l{3pt}r{3pt}){2-3}
Actual classification & Not racist & Racist\\
\midrule
Not racist & 1653 & 1347\\
Racist & 791 & 2209\\
\bottomrule
\end{tabular}
\end{table}

\begin{table}[!h]

\caption{\label{tab:fewshotsingle-sexism}Classification of sexist statements with single-category few-shot learning}
\centering
\fontsize{8}{10}\selectfont
\begin{tabular}[t]{lrr}
\toprule
\multicolumn{1}{c}{ } & \multicolumn{2}{c}{GPT-3 classification} \\
\cmidrule(l{3pt}r{3pt}){2-3}
Actual classification & Not sexist & Sexist\\
\midrule
Not sexist & 2334 & 666\\
Sexist & 1125 & 1875\\
\bottomrule
\end{tabular}
\end{table}

\begin{table}[!h]

\caption{\label{tab:fewshotsingle-hate}Classification of hate speech with single-category few-shot learning}
\centering
\fontsize{8}{10}\selectfont
\begin{tabular}[t]{lrr}
\toprule
\multicolumn{1}{c}{ } & \multicolumn{2}{c}{GPT-3 classification} \\
\cmidrule(l{3pt}r{3pt}){2-3}
Actual classification & Not hate speech & Hate speech\\
\midrule
Not hate speech & 3987 & 2013\\
Hate speech & 1916 & 4084\\
\bottomrule
\end{tabular}
\end{table}

\newpage

\hypertarget{appendxbmuxedubsnorinstruction}{%
\subsection{Few-shot mixed category, without instruction}\label{appendxbmuxedubsnorinstruction}}

\begin{table}[!h]

\caption{\label{tab:fewshotmixed-racism}Classification of racist statements with mixed-category few-shot learning}
\centering
\fontsize{8}{10}\selectfont
\begin{tabular}[t]{llrr}
\toprule
\multicolumn{2}{c}{ } & \multicolumn{2}{c}{GPT-3 classification} \\
\cmidrule(l{3pt}r{3pt}){3-4}
Example set & Actual classification & Not racist & Racist\\
\midrule
1 & Not racist & 107 & 13\\
 & Racist & 5 & \vphantom{1} 55\\
\midrule
2 & Not racist & 102 & 18\\
 & Racist & 9 & 51\\
\midrule
3 & Not racist & 99 & 21\\
 & Racist & 4 & 56\\
\midrule
4 & Not racist & 90 & 30\\
 & Racist & 0 & 60\\
\midrule
5 & Not racist & 92 & 28\\
 & Racist & 3 & \vphantom{1} 57\\
\midrule
6 & Not racist & 94 & 26\\
 & Racist & 2 & \vphantom{1} 58\\
\midrule
7 & Not racist & 84 & 36\\
 & Racist & 1 & 59\\
\midrule
8 & Not racist & 71 & 49\\
 & Racist & 2 & 58\\
\midrule
9 & Not racist & 83 & 37\\
 & Racist & 3 & 57\\
\midrule
10 & Not racist & 108 & 12\\
 & Racist & 5 & 55\\
\midrule
All & Not racist & 930 & 270\\
 & Racist & 34 & 566\\
\bottomrule
\end{tabular}
\end{table}

\begin{table}[!h]

\caption{\label{tab:fewshotmixed-sexism}Classification of sexist statements with mixed-category few-shot learning}
\centering
\fontsize{8}{10}\selectfont
\begin{tabular}[t]{llrr}
\toprule
\multicolumn{2}{c}{ } & \multicolumn{2}{c}{GPT-3 classification} \\
\cmidrule(l{3pt}r{3pt}){3-4}
Example set & Actual classification & Not sexist & Sexist\\
\midrule
1 & Not sexist & 113 & 7\\
 & Sexist & 19 & 41\\
\midrule
2 & Not sexist & 110 & 10\\
 & Sexist & 14 & \vphantom{2} 46\\
\midrule
3 & Not sexist & 110 & 10\\
 & Sexist & 14 & \vphantom{1} 46\\
\midrule
4 & Not sexist & 105 & 15\\
 & Sexist & 12 & \vphantom{1} 48\\
\midrule
5 & Not sexist & 106 & 14\\
 & Sexist & 10 & \vphantom{1} 50\\
\midrule
6 & Not sexist & 103 & 17\\
 & Sexist & 12 & 48\\
\midrule
7 & Not sexist & 110 & 10\\
 & Sexist & 14 & 46\\
\midrule
8 & Not sexist & 100 & 20\\
 & Sexist & 11 & \vphantom{1} 49\\
\midrule
9 & Not sexist & 87 & 33\\
 & Sexist & 11 & 49\\
\midrule
10 & Not sexist & 106 & 14\\
 & Sexist & 10 & 50\\
\midrule
All & Not sexist & 1050 & 150\\
 & Sexist & 127 & 473\\
\bottomrule
\end{tabular}
\end{table}

\newpage

\hypertarget{appendxbmuxedubstryctub}{%
\subsection{Few-shot mixed category, with instruction}\label{appendxbmuxedubstryctub}}

\begin{table}[!h]

\caption{\label{tab:fewshotmixedinstruct-racism}Classification of racist statements with mixed-category few-shot learning, with instruction}
\centering
\fontsize{8}{10}\selectfont
\begin{tabular}[t]{llrr}
\toprule
\multicolumn{2}{c}{ } & \multicolumn{2}{c}{GPT-3 classification} \\
\cmidrule(l{3pt}r{3pt}){3-4}
Example set & Actual classification & Not racist & Racist\\
\midrule
1 & Not racist & 98 & 22\\
 & Racist & 7 & 53\\
\midrule
2 & Not racist & 107 & 13\\
 & Racist & 21 & \vphantom{2} 39\\
\midrule
3 & Not racist & 95 & 25\\
 & Racist & 11 & \vphantom{2} 49\\
\midrule
4 & Not racist & 105 & 15\\
 & Racist & 11 & \vphantom{1} 49\\
\midrule
5 & Not racist & 108 & 12\\
 & Racist & 21 & \vphantom{1} 39\\
\midrule
6 & Not racist & 113 & 7\\
 & Racist & 21 & 39\\
\midrule
7 & Not racist & 110 & 10\\
 & Racist & 11 & 49\\
\midrule
8 & Not racist & 84 & 36\\
 & Racist & 4 & 56\\
\midrule
9 & Not racist & 98 & 22\\
 & Racist & 8 & 52\\
\midrule
10 & Not racist & 71 & 49\\
 & Racist & 1 & 59\\
\midrule
All & Not racist & 989 & 211\\
 & Racist & 116 & 484\\
\bottomrule
\end{tabular}
\end{table}

\begin{table}[!h]

\caption{\label{tab:fewshotmixedinstruct-sexism}Classification of sexist statements with mixed-category few-shot learning, with instruction}
\centering
\fontsize{8}{10}\selectfont
\begin{tabular}[t]{llrr}
\toprule
\multicolumn{2}{c}{ } & \multicolumn{2}{c}{GPT-3 classification} \\
\cmidrule(l{3pt}r{3pt}){3-4}
Example set & Actual classification & Not sexist & Sexist\\
\midrule
1 & Not sexist & 108 & 12\\
 & Sexist & 22 & 38\\
\midrule
2 & Not sexist & 112 & 8\\
 & Sexist & 28 & \vphantom{1} 32\\
\midrule
3 & Not sexist & 116 & 4\\
 & Sexist & 30 & 30\\
\midrule
4 & Not sexist & 112 & 8\\
 & Sexist & 28 & 32\\
\midrule
5 & Not sexist & 116 & 4\\
 & Sexist & 38 & 22\\
\midrule
6 & Not sexist & 116 & 4\\
 & Sexist & 48 & 12\\
\midrule
7 & Not sexist & 111 & 9\\
 & Sexist & 29 & 31\\
\midrule
8 & Not sexist & 114 & 6\\
 & Sexist & 25 & 35\\
\midrule
9 & Not sexist & 116 & 4\\
 & Sexist & 37 & 23\\
\midrule
10 & Not sexist & 109 & 11\\
 & Sexist & 15 & 45\\
\midrule
All & Not sexist & 1130 & 70\\
 & Sexist & 300 & 300\\
\bottomrule
\end{tabular}
\end{table}

\newpage

\hypertarget{appendxc}{%
\section{Effect of misspellings}\label{appendxc}}

\begin{table}[!h]

\caption{\label{tab:profanity}Strings used to extract comments with profanity and edits introduced to test impacts of misspelling and censorship on hate speech classificiation}
\centering
\fontsize{8}{10}\selectfont
\begin{tabular}[t]{ll}
\toprule
Original & Edits\\
\midrule
nigg & n1gg, nig, n*gg, n-gg\\
fuck & fck, phuck, f*ck, f-ck\\
whore & wh0re, whor3, wh0r3, wh*re, wh-re\\
kill & k1ll, kil, k-ll, k*ll\\
rape & r@pe, rap3, r@p3, rap, r*pe, r-pe\\
die & d1e, di3, d13, dye, d*e, d-e\\
bitch & b1tch, bich, bithc, bicht, b*tch, b-tch\\
\bottomrule
\end{tabular}
\end{table}

\begin{table}[!h]

\caption{\label{tab:misspell-results}Zero-shot learning classifications impacted by misspelling and censoring racist comments}
\centering
\fontsize{7.5}{9.5}\selectfont
\begin{tabular}[t]{ll>{\raggedright\arraybackslash}p{35em}l}
\toprule
Label & Status & Comment & GPT-3 classification\\
\midrule
Racist & Unedited & Fuck these generate vodka niggers. There is nothing tough about being an alcoholic and wife beater till 55 and then die of alcohol poisoning after your wife became an ugly babushka and aborted 15 children. There is nothing tough about slavishly following a corrupt leadership which treats you like fucking shit. & Yes\\
Racist & Edited & fck these generate vodka niggers. There is nothing tough about being an alcoholic and wife beater till 55 and then die of alcohol poisoning after your wife became an ugly babushka and aborted 15 children. There is nothing tough about slavishly following a corrupt leadership which treats you like fcking shit. & No\\
Racist & Edited & phuck these generate vodka niggers. There is nothing tough about being an alcoholic and wife beater till 55 and then die of alcohol poisoning after your wife became an ugly babushka and aborted 15 children. There is nothing tough about slavishly following a corrupt leadership which treats you like phucking shit. & No\\
Racist & Unedited & Fuck off niggers WV is full & Yes\\
Racist & Edited & phuck off niggers WV is full & No\\
Racist & Unedited & Why are niggers superior athletically to all other races? & Yes\\
Racist & Edited & Why are n*ggers superior athletically to all other races? & No\\
Racist & Unedited & Stand the fuck up and take them out all you fucking lazy white males! fuck being called a racist protect your woman and children. If you find it in the street kill it and bury it right there these animals will wipe you out if your not careful & No\\
Racist & Edited & Stand the phuck up and take them out all you phucking lazy white males! phuck being called a racist protect your woman and children. If you find it in the street kill it and bury it right there these animals will wipe you out if your not careful & Yes\\
Racist & Unedited & Ima fucking stab u up bruv. Fucking white prick bruv.. fucking racist.  Gammon & Yes\\
Racist & Edited & Ima fcking stab u up bruv. fcking white prick bruv.. fcking racist.  Gammon & No\\
Racist & Edited & Ima f*cking stab u up bruv. f*cking white prick bruv.. f*cking racist.  Gammon & No\\
Racist & Unedited & fucking monkeys kill them all & No\\
Racist & Edited & fucking monkeys k-ll them all & Yes\\
\bottomrule
\end{tabular}
\end{table}

\newpage

\newpage
\singlespacing 
\renewcommand\refname{References}
\bibliography{references.bib}

\end{document}